# Implementation of perception and action at nanoscale

Sylvain Marlière[‡+], Jean-Loup Florens[‡], Florence Marchi[*†], Annie Luciani[‡], Joël Chevrier[*†]
(‡) Laboratoire ICA, INPG, 46 Avenue Félix Viallet, 38000 Grenoble
(+)ERGOS-Technologies, INPG, 46 Avenue Félix Viallet, 38000 Grenoble
(*)Institut Néel, CNRS/UJF, 25 Avenue des Martyrs, BP 166, 38042 Grenoble cedex 9
(†)Université Joseph Fourier - BP 53 - 38041 Grenoble Cedex 9
E-mail: joel.chevrier@grenoble.cnrs.fr, luciani@imag.fr

**Abstract**

*Real time combination of nanosensors and nanoactuators with virtual reality environment and multisensorial interfaces enable us to efficiently act and perceive at nanoscale. Advanced manipulation of nanoobjects and new strategies for scientific education are the key motivations. We have no existing intuitive representation of the nanoworld ruled by laws foreign to our experience. A central challenge is then the construction of nanoworld simulacrum that we can start to visit and to explore. In this nanoworld simulacrum, object identifications will be based on probed entity physical and chemical intrinsic properties, on their interactions with sensors and on the final choices made in building a multisensorial interface so that these objects become coherent elements of the human sphere of action and perception. Here we describe a 1D virtual nanomanipulator, part of the Cité des Sciences EXPO NANO in Paris, that is the first realization based on this program.*

## 1. Introduction

Nanotechnologies are already involved in cellular phones or laptops. This is nanoelectronic. This provides us with tools for massive and easy real time treatment of informations. Consequences are a new representation of space and time and revolutionized scheme of interactions among people. If ubiquity is the permanent possibility to have simultaneous interactions with different people in different places, then this is provided by nanotechnologies. Nanotechnologies will quite soon make more available. Network of nanosensors and nanoactuators will significantly extend our capacity to experience the world around us. The nanonose [1] should routinely enhance our capacity to detect different chemical species at level down to few molecules if not a single one. Medical applications such as immediate diagnostics are clearly numerous [2]. These new ways of treating information and of interacting with the matter either living or inert can be implemented in different ways using nanotechnologies. Tangible media strategies for example explore the possibility to inscribe bits of information in objects around us [3]. Our ability to physically interact with an environment is then essentially left unchanged. It is the information content of this environment that is broadly enhanced. A very different strategy, albeit not concurrent, is to extend in real time our perception and actions in the world around us [4-|6]. It is then here not reality that is augmented, but our ability to directly experience the world around us. New perception and action modalities are technologically built, based on a detailed control of how new aspects of matter organization can interact with us. Open area are numerous and are not limited to the nanoworld which however is the most challenging example. It can be amplification of minute morphological changes revealing to our senses how much grounds, houses,... are constantly vibrating, moving. One can realize how much water no longer resembles the liquid we are used to if we consider it at the micrometer scale or how much living microorganism constitution and behavior are constrained by rules that are foreign to our daily experience. One way to develop this extension of the sphere where our senses are efficient can be based on real nanosensors and nanoactuators. An another approach is to use virtual environments which can offer the nanoworld to us through real time multisensorial interfaces. This can dramatically enhance possibilities for easy exploration of remote realities foreign to our senses and can trigger a spontaneous motivation of the user similar to the one observed in video game player. Interest in these developments is for us threefold. The first two aspects are the application fields we have chosen: nanomanipulation and elementary scientific education. The third one relates to a fundamental aspect of this program and roots the chosen applications. It is a major question: if the nanoworld is our chosen playground, shall we be able to define a multisensorial interface that once implemented in the nanoscene both preserves the original specificities of the nanoworld and develops the trainee ability to master this interface for an interesting and efficient use. To be relevant and successful, this program should be structured by interdisciplinary developments. The first aforementioned aspect is related to mechanical nanomanipulation of real nanoobjects. Hand on control combined with the use of sophisticated sensors and actuators is required. It is still a research project. The second aspect is more focused on new strategies for scientific education. Real time use of senses to explore "new worlds" thanks to new virtual environments can reveal aspects of matter behavior to anybody that are classically approached only through advanced scientific education.

A central part of this program is not a new idea. It goes back to birth of experimental science with use of telescope by Galileo to observe the Moon and to come to the immediate conclusion that the Moon is Earth like. As immediately emphasized by Galileo, this dramatic change in the human representation of the Universe is caused by direct use of senses technically extended by an instrument and not by a posteriori rational demonstration.

Our proposal can be seen as a revival of this famous tale. There is however a major difference. Two questions can illustrate the need for new approaches based on implementation of nanoscene in virtual environments. As nanoscale is gradually approached, continuous description no longer stands and the molecular discontinuous structure of matter is revealed. Atomic scale is a radical rupture with our common experience that is based on the objective existence of isolated continuous objects. Second is: can we offer ourselves the possibility to see and to touch an electron, a particle that has a mass and an electric charge but has no classical material spatial extension in the sense of a material sphere, although severely constrained by the Pauli principle in its collective spatial properties that are at the root of stability of matter stability. In fact seeing and touching an electron has no intrinsic meaning. Electron based objects can however be created and our interaction with these unusual objects defined. Shall we be able to build a bridge between our perception and "electron based objects" ?

If connection to the real world is considered, the simulacrum involves a numerical modelisation of the nanoworld. All measures transferred by nanosensors are translated trough real time simulation to this nanoscene, where they are experienced by the users. Symmetrically acts done by the user are both analyzed through their effects in the nanoscene where they can be simultaneously perceived for immediate feedback and transferred to the real scene for direct manipulation.

This approach is summed up in figure 1.

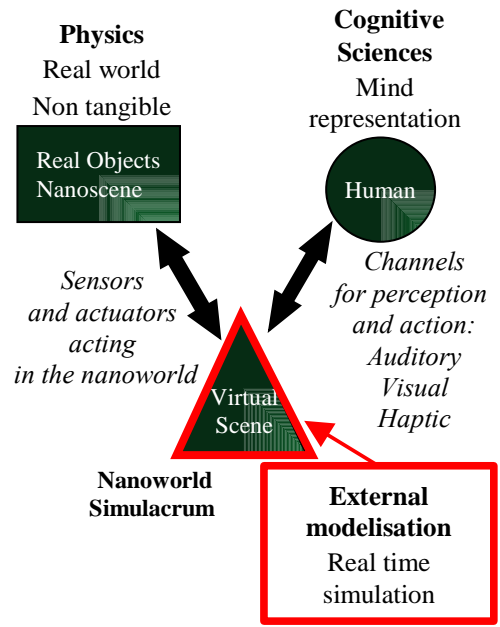

Figure 1: Basics of real time extension of perception and action to non tangible objects and foreign aspects of matter for our senses. Schematic description.

## 2. Feeling the nanoworld

Our point is then that seeing, hearing and touching can be technically transferred to the nanoscale. We are now testing the efficiency of this idea in scientific exhibitions. Through the use of nanosensors and nanoactuators, real time processing and visual, auditory and haptic interfaces, a multisensorial reconstruction of the nanoworld can be proposed so that one has the feeling to directly explore a foreign space. As a first step, transfer of direct repulsive contact has already been realized by many groups. Although very useful, nanomanipulators based on this, do not really challenge our perception. In this case, for example, the continuous and geometrical description of a carbon nanotube makes it appearing like a spaghetti in a plate. That we can use such a straightforward comparison is meaningful. Direct contact is finally related to the impenetrability: two solids cannot occupy the same place. They must be separated. The direct contact we are used to at our scale occurs on a distance that is finally nanometric.

Among many specific aspects, we have first chosen the universal long range interaction between nanobjects. In real life, we are not sucked by walls as we are moving close to them. At nanoscale, this is the case. This effect is often quoted as one of the most important atomic properties to understand matter structure. It immediately prevents us to define nanoobjects by intrinsic sharp geometrical contours as it is mostly always the case at our scale. Control of interaction between sensors and nanoobjects necessarily enters perception and representation. Again how can we see, touch and even manipulate with hands, for example the electron droplets created by light in CCD sensors?

## 3. A virtual experience : the nanocontact

The most obvious and simple way to experience interaction with a table using an instrument is to touch its surface using a stick. No need to explain the user in detail what is the situation: the user has a stick in hand and he uses it to touch the table surface coming from above. This is obvious to everybody as is then the dramatic change at nanoscale in such a simple situation introduced by the long range interaction. This is what has been proposed to a general audience at public exhibition, first at the CCSTI Grenoble (Exposition Nanotechnologies) and at the Cité des Sciences in Paris (Expo Nano) and now at the Globe CERN Geneva. The stick, table surface and their interaction are numerically modeled (see figure 2) first at our scale so that, as a reference, one has the experience of the direct contact then at the nanoscale where this simple experience in fact exactly reproduces the so called force approach curves obtained using an Atomic Force Microscope.

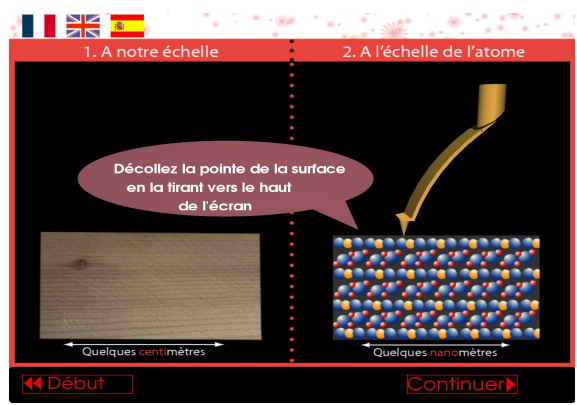

Figure 2: Copy of the screen used during the manipulation. Left: wood table at meter scale; right: atomic layer here directly touched by the manipulated elastic stick. One can continuously move from right to left and back.

One major point in this simulation is the insertion of a virtual stick that has the properties of a real one between the simulated nanoscene and the user. This stick is elastic with a major consequence as the stick end (the tip in figure 2) is getting too close to the surface, it is irreversibly sucked by the surface. It is for the user a challenge or a game very close to the actual experimental situation to stabilize the tip as close as possible to the surface without touching it. To insert a realistic tool has been done on purpose. First it carries the fact that we are not implementing perception and action in an ideal world as described by physics even though it is here a virtual world. However it must be clear that the modeled instrument characteristics, here the stick, are not due to limitations of the haptic interface. To the contrary this can be done thanks to high performance (force amplification, positioning precision, large bandwidth) of the ERGOS haptic interface.

This virtual Atomic Force Microscope is coupled to this advanced haptic interface and a sonification and visualization system (see figure 3).

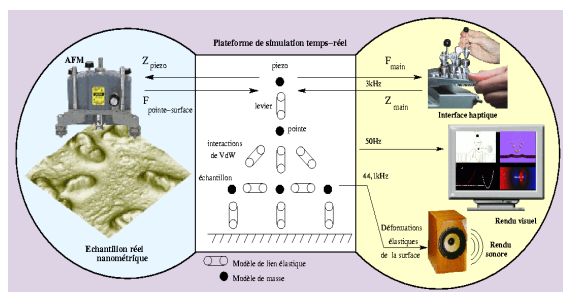

Figure 3: Image of the virtual machine. Left part: image of a real AFM; Center: elastic model of the stick, table and interaction based on mass and spring combination.

Using this machine, as expected, everybody can then compare his feelings when directly experiencing characteristics of the usual contact at our scale or specificities of the nanocontact.

We then believe we have here produced a set up that leads anybody in a few minutes to directly experience the long range interaction at nanoscale in his hands. This is finally reminiscent of the Galileo procedure: after manipulation one is supposed to be convinced about the existence and the importance of this universal long range interaction.

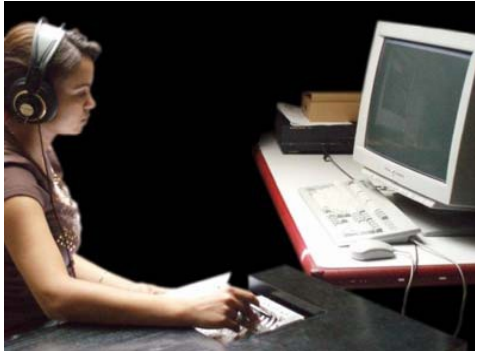

Figure 4: Manipulation using the virtual AFM we have built. In her hand, the haptic interface lever whose details are shown in figure 5.

One should suddenly realize the huge change it would be to invest such a world based on rules that are totally remote to our common experience in the daily environment.

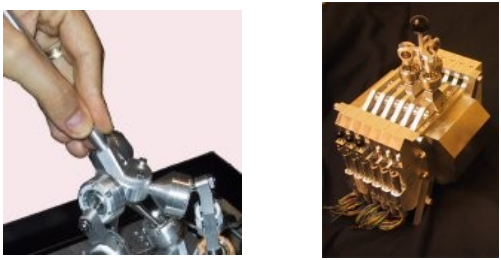

Figure 5: Details of the modular haptic interface here in a 3D configuration.

## 4. A preliminary evaluation

This comment is in fact our expectation. Although we have qualitatively observed this is what happening with users, this remains a weak conclusion. In order to somewhat strengthen this analysis, using inserted counters, we have analyzed the time spent by visitors on this machine at Expo Nano in La Cité des Sciences Paris.

From the March 19th 2007 to the June 1st 2007, visitor number has been 5749 i.e. about 90 visitors a day during 5 to 6 hours. The overall time on the set up spent by a user has been about 4 minutes and the manipulation time per user close to 2 minutes. From these numbers we can draw some first conclusions. This machine was offered to the public which used it freely. 90 users a day is a large number. Second an average time of 4 minutes including 2 minutes spent using simulations, appears to us reasonable to enter the system, understand the problem and get the prepared message. This result is consistent with our expextation.

This experience has put in a large audience hands a robotic set up for an extended period of time. It is then a first and encouraging step in the development of real time virtual systems equipped with multisensorial interfaces that enable one to discover aspects of our environment well scientifically described but that are totally remote to our senses. It is a research subject to observe the way people discover this surprising piece of reality and how they can get trained to it. How this training is influenced by the definition of the multisensory interface and the combination of sensory channels will be a key question. The a priori level of user scientific education needed to master the system depending on its configuration will also be interesting especially to appreciate the relevance of this procedure in scientific education.